\documentclass[journal]{IEEEtran}
\usepackage{cite}
\usepackage{amsmath,amssymb,amsfonts}
\usepackage{algorithmic}
\usepackage{array}
\usepackage{graphicx}
\usepackage{textcomp}
\usepackage{xcolor}
\usepackage{bm}
\usepackage{graphicx} 
\usepackage{epstopdf}
\usepackage[normalem]{ulem}
\useunder{\uline}{\ul}{}
\usepackage{multirow}
\usepackage{booktabs}
\usepackage{url}
\usepackage{hyperref}
\usepackage{color}
\usepackage{subfigure}
\usepackage{amsmath}
\usepackage{multicol}
\usepackage{multirow}
\usepackage{booktabs}
\usepackage{bbding}
\usepackage{longtable}
\usepackage{supertabular}
\usepackage{pifont}
\usepackage{float}
\hypersetup{
	colorlinks=true,
	linkcolor=blue,
	filecolor=purple,      
	urlcolor=blue,
	citecolor=green,
}

\ifCLASSINFOpdf
\else
\fi

\begin{document}
%
\title{Leveraging Multi-stream Information Fusion for Trajectory Prediction in Low-illumination Scenarios: A Multi-channel Graph Convolutional Approach}
\author{Hailong Gong\textsuperscript{1, *},
        Zirui Li\textsuperscript{1, 2, *},~\IEEEmembership{Graduate Student Member,~IEEE},
        Chao Lu\textsuperscript{1},~\IEEEmembership{Member,~IEEE}, \\
        Guodong Du\textsuperscript{3}
        and Jianwei Gong\textsuperscript{1},~\IEEEmembership{Member,~IEEE}
\thanks{This work is supported by the National Natural Science Foundation of China under Grants 61703041 and U19A2083, and is also supported by China Scholarship Council (CSC).}
\thanks{$^{*}$Hailong Gong and Zirui Li contribute equally in this work.}  
\thanks{$^{1}$Hailong Gong, Zirui Li, Chao Lu and Jianwei Gong are with the School of Mechanical Engineering, Beijing Institute of Technology, Beijing 100081, China. (Email\tt\small: gonghailong2021@163.com; 3120195255@bit.edu.cn; chaolu@bit.edu.cn;  gongjianwei@bit.edu.cn)}
\thanks{$^{2}$Zirui Li is also with the Chair of Traffic Process Automation, "Friedrich List" Faculty of Transport and Traffic Sciences, TU Dresden, Germany.}
\thanks{$^{3}$Guodong Du is with Institute of Dynamic System and Control, ETH Zurich, Sonneggstrasse 3, 8092 Zurich, Switzerland. (Email\tt\small: guoddu@ethz.ch)}
\thanks{(corresponding authors: Chao Lu, Jianwei Gong)}
}

\markboth{IEEE Transactions on Intelligent Transportation Systems}%
{Shell \MakeLowercase{\textit{et al.}}: Bare Demo of IEEEtran.cls for IEEE Journals}
\maketitle
\begin{abstract}
Trajectory prediction is a fundamental problem and challenge for autonomous vehicles. Early works mainly focused on designing complicated architectures for deep-learning-based prediction models in normal-illumination environments, which fail in dealing with low-light conditions. This paper proposes a novel approach for trajectory prediction in low-illumination scenarios by leveraging multi-stream information fusion, which flexibly integrates image, optical flow, and object trajectory information. The image channel employs Convolutional Neural Network (CNN) and Long Short-term Memory (LSTM) networks to extract temporal information from the camera. The optical flow channel is applied to capture the pattern of relative motion between adjacent camera frames and modelled by Spatial-Temporal Graph Convolutional Network (ST-GCN). The trajectory channel is used to recognize high-level interactions between vehicles. Finally, information from all the three channels is effectively fused in the prediction module to generate future trajectories of surrounding vehicles in low-illumination conditions. The proposed multi-channel graph convolutional approach is validated on HEV-I and newly generated Dark-HEV-I, egocentric vision datasets that primarily focus on urban intersection scenarios. The results demonstrate that our method outperforms the baselines, in standard and low-illumination scenarios. Additionally, our approach is generic and applicable to scenarios with different types of perception data. The source code of the proposed approach is available at \href{https://github.com/TommyGong08/MSIF}{https://github.com/TommyGong08/MSIF}.
\end{abstract}
\IEEEpeerreviewmaketitle
\section{Introduction}
With the rapid development of autonomous driving, it has become apparent that ensuring the safety of autonomous systems in traffic scenarios is a necessary condition for the widespread adoption of autonomous driving\cite{koopman2017autonomous}. For self-driving cars to have driving capabilities comparable to those of human drivers, it is essential to understand the state of surrounding vehicles and predict their trajectories\cite{schwarting2018planning}.

\subsection{Motivation}
Trajectory prediction is the key technology that contributes to the development of autonomous driving, and it has become a popular topic of research in recent years\cite{gulzar2021survey}. Its essence is to infer possible intentions and future trajectories for agents based on their historical trajectories and environmental context, which include camera and lidar-based sensors.  However, most of the available computer vision technologies are based on visible light cameras, and thus can only be used under the condition of normal light and clear weather~\cite{chen2017error}. Previous studies on trajectory prediction focused on scenarios in normal-illumination environments with standard lights and often failed in low-light conditions, which makes most of the state-of-the-art models not suitable at night~\cite{li2021deep}. The survey in~\cite{NHTSA} shows that 51\% of fatal traffic accidents in U.S. at nighttime, especially in rural areas with extremely low-illumination. Hence, accurate trajectory prediction in low-illumination scenarios is crucial for the traffic safety.

\subsection{Related Works} 
Numerous types of research have proposed various methods for predicting trajectories. Physics-based methods employ the dynamics or kinematics models of vehicles \cite{huang2022survey}. In most cases, a simple physics model is preferred because complex physics models provide only marginal improvements in predictive accuracy. Kalman filtering is popularly applied in physics-based methods. \cite{kaempchen2004imm} models the noise of the current state of vehicles using Kalman filtering techniques. Based on Gaussian Mixture Model(GMM), \cite{li2018development} predicts the lookahead distance for the autonomous vehicles and \cite{li2022personalized} successfully recognizes the braking intensity levels of drivers. Combining vehicle-to-vehicle (V2V) communication and the Kalman filter, \cite{zhang2017method} predicts ego-vehicle trajectories to avoid obstacles. However, the accuracy of physics-based methods is heavily dependent on the description of physics models. If dynamic models of vehicles change, physics-based methods can only provide short-term predictions.

Recently, the most popular methods for predicting trajectories are based on deep learning, as they can effectively integrate physical constraints, interactions, and scene understanding \cite{huang2022survey}. To solve the problem of insufficient data and improve modeling efficiency, transfer learning is used in driver behavior modeling\cite{lu2019virtual}, especially in the lane-changing scenario \cite{lu2019transfer,li2019transferable,li2020importance,gong2019comparative}. \cite{xu2021tra2tra} extracts features of spatial interactions with attention mechanism and employs the LSTM network to determine their temporal dependence. \cite{gupta2018social} presents a social generative adversarial network (GAN) model that focuses on the normalization and rationality of trajectories.
To effectively capture the social behaviors of relevant pedestrians, a graph neural network is implemented in social behavior modeling and trajectory prediction based on their timely location and speed direction \cite{li2021hierarchical, li2021interactive}. Graph neural network employs graphs to represent traffic scenarios in which nodes represent vehicles and edges represent the degree of interaction between vehicles \cite{zhang2019stochastic, yan2018spatial,haddad2019situation,mohamed2020social}. \cite{yan2018spatial} constructs the spatial temporal graph model, in which the temporal graph extracts personal information and the spatial graph extracts pedestrian interaction information. The Social-STGCNN model proposed by \cite{mohamed2020social} employs a graph convolutional neural network to embed the spatial temporal graph and a time extrapolator to determine trajectories. 

However, the physic model-based and deep learning-based trajectory prediction methods mainly focus on normal driving conditions. As illumination conditions change throughout the day, it is necessary for self-driving vehicles to make image enhancement and extract scene feature in the low-brightness environments \cite{li2021deep}. \cite{fu2016fusion} develops a fusion-based enhancing method for weakly illuminated images. \cite{loh2019getting} proposes a dataset with low-light images, finding that the effects of low-light reach far deeper into the features than can be solved by simple "illumination invariance". \cite{li2021deep} develops am image enhancement approach for autonomous driving at night. Optical flow information is introduced to ensure the consistency of transformed brightness or to realize optical flow tracking in response to the difficulty posed by low-illumination conditions. \cite{jung2020multi} uses optical flow to improve image quality in low-light conditions. Utilizing dense optical flow, \cite{pikoulis2021leveraging} encodes motion between consecutive frames and achieves visual emotion recognition in low resolution and poor illumination. \cite{rashed2019fusemodnet} integrates optical flow and LiDAR perception data for moving object detection in autonomous driving under low-light conditions. The above works have investigated image enhancement and object detection in low-light conditions of autonomous driving, but little research has explored trajectory prediction in low-illumination environment.

A series of trajectory prediction approaches, such as physics-based, deep learning methods, have achieved state-of-the-art performance for normal driving conditions, but for complex autonomous driving traffic scenarios, environmental conditions are not constant. Autonomous vehicles can be exposed to extreme conditions such as low and strong lights. Moreover, some research work related to computer vision on low-light conditions seldom pay attention to the trajectory prediction problem \cite{jung2020multi,pikoulis2021leveraging, rashed2019fusemodnet}, making it difficult to solve the trajectory prediction problem for extreme conditions.
\subsection{Contributions}
\begin{figure*}[htbp]
\centering
\includegraphics[width=0.8\linewidth, height=0.75\textwidth]{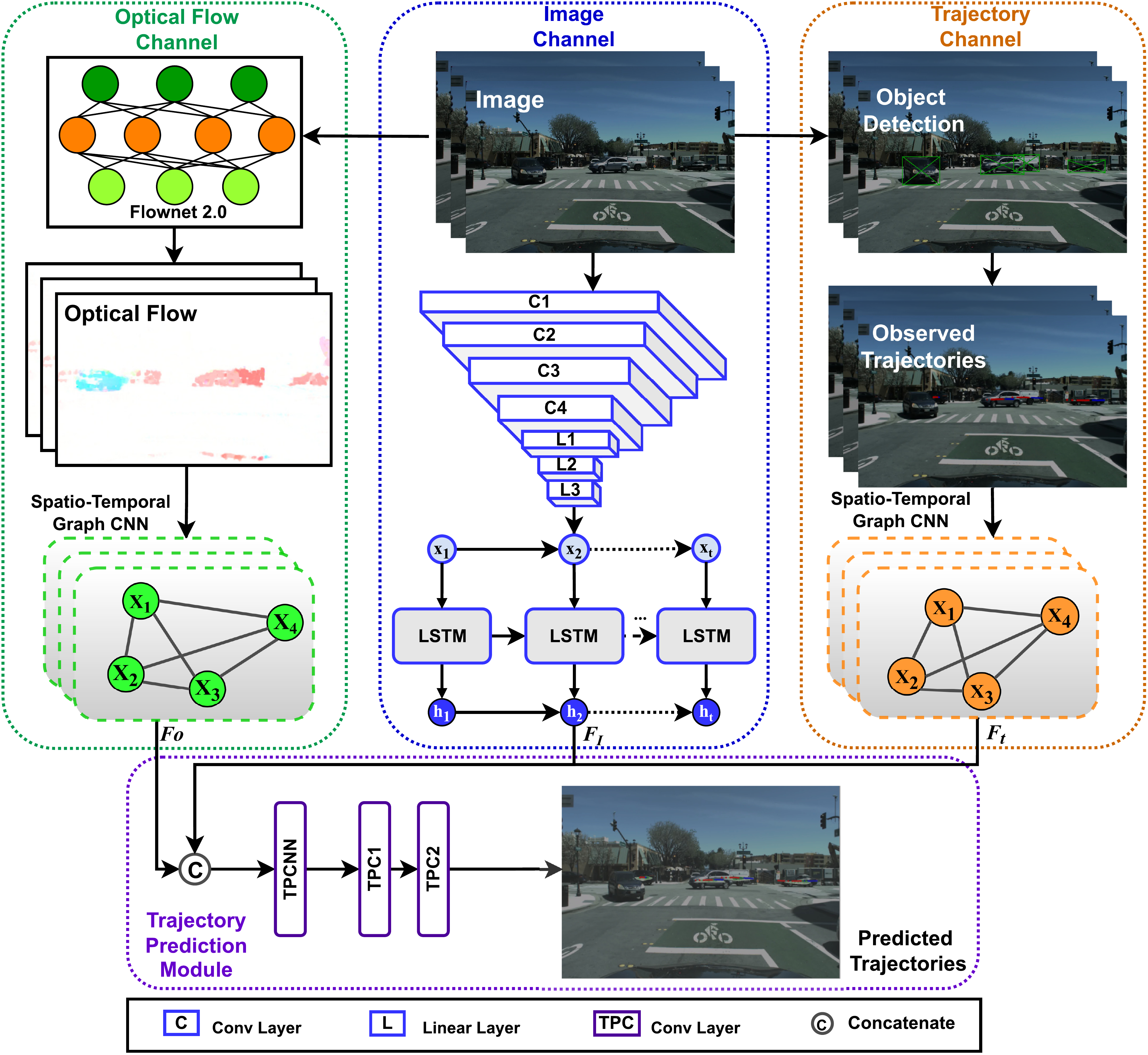}
\caption{ The model consists of three channels: optical flow, image, and trajectory (from top left to top right). The output of this model is the distribution of predicted trajectories. The proposed approach implements CNN and LSTM layers for feature extraction and scene understanding for image information. A Spatio-temporal graph convolutional neural network is used for feature extraction for optical flow and trajectory information. All the extracted features will be concatenated and transferred into the trajectory prediction module. Considering the weight and efficiency of the model, a convolution neural network is adopted in the future trajectories prediction module.}
\label{framework}
\end{figure*}
To overcome the detrimental effects of low-light conditions on the autonomous driving, especially in the trajectory problem, this research proposes a multi-stream information (heterogeneous data) fusion-based method, MSIF, for trajectory prediction in low-illumination scenarios. The proposed method combines trajectories, optical flow, and image information, which ensures adaptability to various luminance levels and especially overcomes low-illumination conditions. To model the interaction in low-brightness conditions for trajectory prediction, the graph convolutional neural network (GCN) is applied to represent spatial-temporal features of trajectories \cite{mohamed2020social} and instantaneous speed of surrounding vehicles (from optical flow). Meanwhile, local spatial differences are identified using a novel recurrent-based image feature extraction technique \cite{d2021network}. To simulate realistic low-light driving conditions, the Dark-HEV-I dataset is derived from the HEV-I dataset by adjusting the image brightness. The main contributions of this paper are summarized as follows:
    \subsubsection{} A novel trajectory prediction method is proposed for low-illumination conditions by leveraging multi-stream information fusion, which flexibly integrates image, optical flow, and object trajectory information. The proposed method designs the ST-GCN-based method for temporal and spatial information representation and incorporates a novel multi-stream information fusion mechanism into its architecture.
    \subsubsection{} To simulate low-illumination driving conditions and evaluate the effectiveness of the proposed method, the Dark-HEV-I dataset is derived from the HEV-I dataset with the same scale. In Dark-HEV-I dataset, the low-illumination images are generated by adjusting the exposure of the original images, and optical flow is produced by using the low-illuminated images. Experimental results demonstrate that the proposed method could maintain high performance in the Dark-HEV-I dataset.
\subsection{Outline}
This paper is organized as follows. In section~\ref{Problem Formulation}, the formulation of the multi-stream trajectory prediction is detailed. Section~\ref{Methodology} begins with a description of the proposed method, followed by the formulation of graph representation, image feature extraction, and information fusion. Section~\ref{Experiments} shows the HEV-I dataset, the newly generated Dark-HEV-I dataset, implementation details and experimental results. Finally, the conclusion is presented in Section~\ref{Conclusion}.
\section{Problem Formulation} \label{Problem Formulation}
As shown in Fig.~\ref{framework}, a novel method involving an image channel, optical flow channel, trajectory channel, and the fusion module of trajectory prediction is proposed to predict trajectories of moving objects in the field of view. This study assumes that an autonomous driving vehicle can obtain heterogeneous data from its sensors. At each time step, at least the front-view image is provided. The speed and direction of objects in motion are not provided. As front-view camera images are commonly used in autonomous vehicles, this study assumes that the optical flow can be generated from the original images. Historical trajectories of objects in the scenario are obtained through object detection and trajectory tracking. Specifically, the model's input at time $t$ is defined as follows:
\begin{equation}
    \bm S_t=\{\bm I_t, \bm O_t, \bm X_t\}
\end{equation}
\begin{equation}
    \bm I_t=[\bm{M}_{t-t_{obs}}, ..., \bm{M}_{t-1}, \bm{M}_{t}]
\end{equation}
where $\bm I_t$ represents the image sequence captured by the front-view camera on the automatic vehicle. $\bm O_t$ and $\bm X_t$ denote the optical flow sequence and the sequence of the set formed by objects' trajectories in each frame, respectively, where
\begin{equation}
    \bm O_t=[\bm{J}_{t-t_{obs}}, ..., \bm{J}_{t-1}, \bm{J}_{t}]
\end{equation}
\begin{equation}
    \bm X_t=[\bm{P}_{t-t_{obs}}, ..., \bm{P}_{t-1}, \bm{P}_{t}]
\end{equation}
where $\bm{P}_{t}$ is the set of the objects' trajectories at time $t$, $\bm{P}_{t} = \{ (x_{t}^{i}, y_{t}^{i}) \ | \ i = 0, 1, ..., n\}$, where $n$ is a variable parameter due to the change of the objects number in the view at different times.

Furthermore, the trajectory and optical flow information will be represented by graphs, in which each node stands for a vehicle. At each time step $t$, the  information of each $\bm{M}_{t}$ is abstracted into a graph $G_t=(V_t, A_t)$, in which $V = \{ v^i \ | \  \forall i \in {1,2,...,N}\}$. $N$ is the total number of objects that appear in that sequence. $V_t$ represents the node, and attributes of $v_i$ are the coordinate of each objects in pixel coordinates, denoted as $v_i = (x_i, y_i)$. $A_t$ represents the adjacency matrix and $A_t = \{ a_t^{i,j} \ | \ \forall i, j \in 1,2,...,N  \}$. For the same sequence, $A_t$ weights vertices' contributions to each in the convolution operation. Thus, a kernel function can be considered as prior knowledge about the interactive degree between vehicles as it maps attributes at $v_i$ and $v_j$ at time step $t$ to the value $a_t^{i,j}$.
\begin{figure}[thbp]
\centering
\includegraphics[width=0.9\linewidth,height=0.45\textwidth]{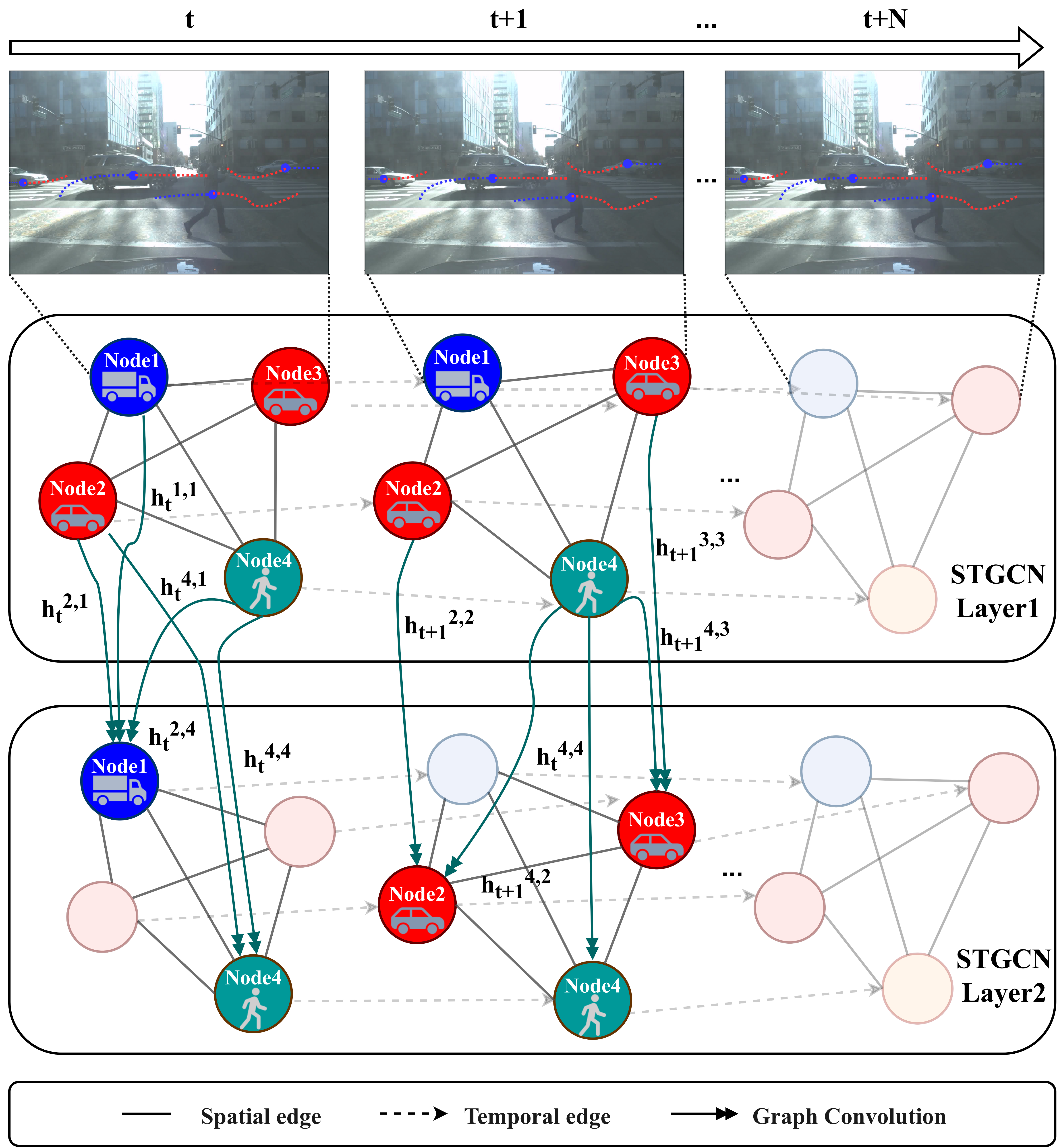}
\caption{ This figure presents the process of graph convolution embedding. The spatial feature forward from node $i$ to node $j$ at time step $t$ is denoted as $h_t^{i, j}$ used in graph convolution. During the graph convolution, the spatial features will be taken part in the operation.}
\label{STGCNN}
\end{figure}
As marked in Fig. \ref{STGCNN}, the spatial feature forwarded from node $i$ to node $j$ at time step $t$ is denoted as $h_t^{i, j}$, used in graph convolution. During the graph convolution, spatial features will be taken part in the operation. Given the same definitions of the sampling function and weight function in \cite{yan2018spatial}, the graph spatial convolution is formulated as:
\begin{equation}
    f_{out}(v_{ti}) = \sum \limits_{v_{tj} \in B(v_{ti})} \frac{1}{Z_{ti}(v_{tj})} f_{in}(v_{tj}) \cdot \bm{W}(l_{ti}(v_{tj}))
\end{equation}
where $B(v_{ti})$ represents the neighbor set of $v_{ti}$, and $Z_{ti}(v_{tj})$ is the normalizing term.

Predicted trajectories are assumed to follow the bi-variate Gaussian distribution, which is estimated to $ \Omega(\hat{\mu}, \hat{\sigma}$, and $ \hat{\rho})$. The output of the model at time $t$ is defined as $\bm R_t=[\bm{P}_{t+1}, \bm{P}_{t+2}, ..., \bm{P}_{t+t_{pred}}]$. Defining parameters of the $i$-th object's bi-variate Gaussian distribution at the moment $t$ as $\mu_t^i$,$\sigma_t^i$, $\rho_t^i$, the output can  be formulated as:
\begin{equation}
    \bm{\Psi}(\bm R_{t}^i \ | \ \bm S_{t}) \sim \Omega(\hat{\mu}_{t}^i, \hat{\sigma}_{t}^i, \hat{\rho}_{t}^i), i=1, 2, ..., n
\end{equation}
The ground truth of predicted trajectories is denoted as $Y_{t+1:t+t_{pred}}$. The goal of the proposed approach is to precisely map observations to predictions, which can be formulated as:
\begin{equation}
    \arg\min\limits_{f^{}\in \bm{F}} \bm{L}(Y_{t+1:t+t_\text{pred}}, f^{}(S_{t}))
\end{equation}
where $\bm{F}$ is the model set in the training process, $f$ represents the model, and $\bm{L}$ is the measure of prediction error.
\section{Methodology} \label{Methodology}
This section begins with a summary of the methodology, including model inputs, outputs, data flows, and structural characteristics, as shown in Fig. \ref{framework}. Then, the principle and functions of three channels in the proposed model are described: the optical flow channel, the trajectory channel, and the image channel. Finally, the last subsection introduces the trajectory prediction module, specifically describing information fusion methods.
\subsection{Multi-stream Information Fusion Framework}
The low illumination environment poses the challenge for trajectory prediction, which threatens the safety of autonomous vehicles. Due to the insufficient light and inadequate understanding of the scenario, it is inconsiderate to merely use trajectory information for prediction. For trajectory prediction in the low illumination scenarios, it is intuitive that the prediction method should make an image enhancement and capture the information about vehicles in motion. Therefore, our approach innovatively utilizes optical flow and image information for trajectory prediction besides trajectory information.

As shown in Fig. \ref{framework}, the MSIF method combines front-view image streams, optical flow, and trajectories from object detection using three input channels: 1) Optical channel, 2) Image channel, and 3) Trajectory channel. The model's inputs consist of the pixel matrices of the image stream, the optical flow, and the trajectories, where the image stream is the original data, and the optical flow and trajectories are generated from the image stream.

For the image channel, after obtaining the images of the front-view camera, the approach first resizes the images. It implements the Convolutional Neural Network (CNN) and Long Short-Term Memory (LSTM) layers to extract the features of the resized image. Several optical flow generation techniques, such as \cite{ilg2017flownet, dosovitskiy2015flownet}, have been developed for the optical flow channel. Flownet 2.0 generates optical flow in this framework\cite{ilg2017flownet}. Then, the optical flow channel distinguishes moving objects from background and enhances the image channel by leveraging a Spatio-temporal graph convolutional neural network. For the trajectory channel, the target detection algorithm takes original images as inputs and generates trajectories of the geometric center of the bounding boxes. Trajectories are served as inputs of a Spatio-temporal graph convolutional neural network for interactive behavior modeling in the trajectory channel. Finally, the trajectory prediction module accepts extracted features from the three channels mentioned above, where the features are fused and used to generate predicted trajectories. Considering the weight and efficiency of the model, a convolution neural network is utilized in the future trajectory prediction module.

The output of this model is the bivariate Gaussian distribution of the predicted trajectories. It must be emphasized that all trajectory coordinates in this paper are in the pixel coordinate system.
\begin{figure}[]
\centering
\subfigure[]{
\begin{minipage}[t]{\linewidth}
\centering
\includegraphics[width=0.8\linewidth]{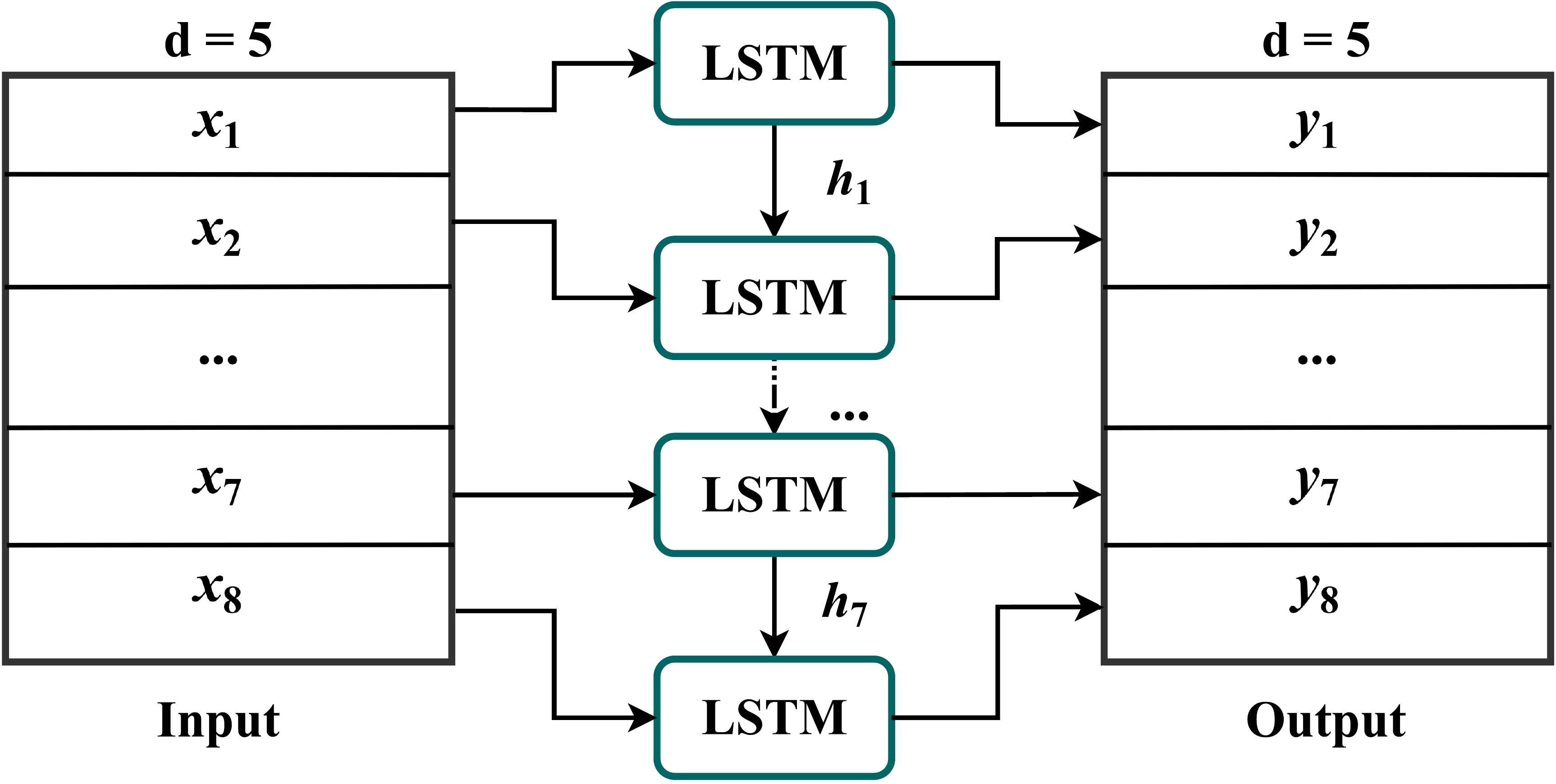}
\end{minipage}%
}%

\subfigure[]{
\begin{minipage}[t]{\linewidth}
\centering
\includegraphics[width=0.7\linewidth]{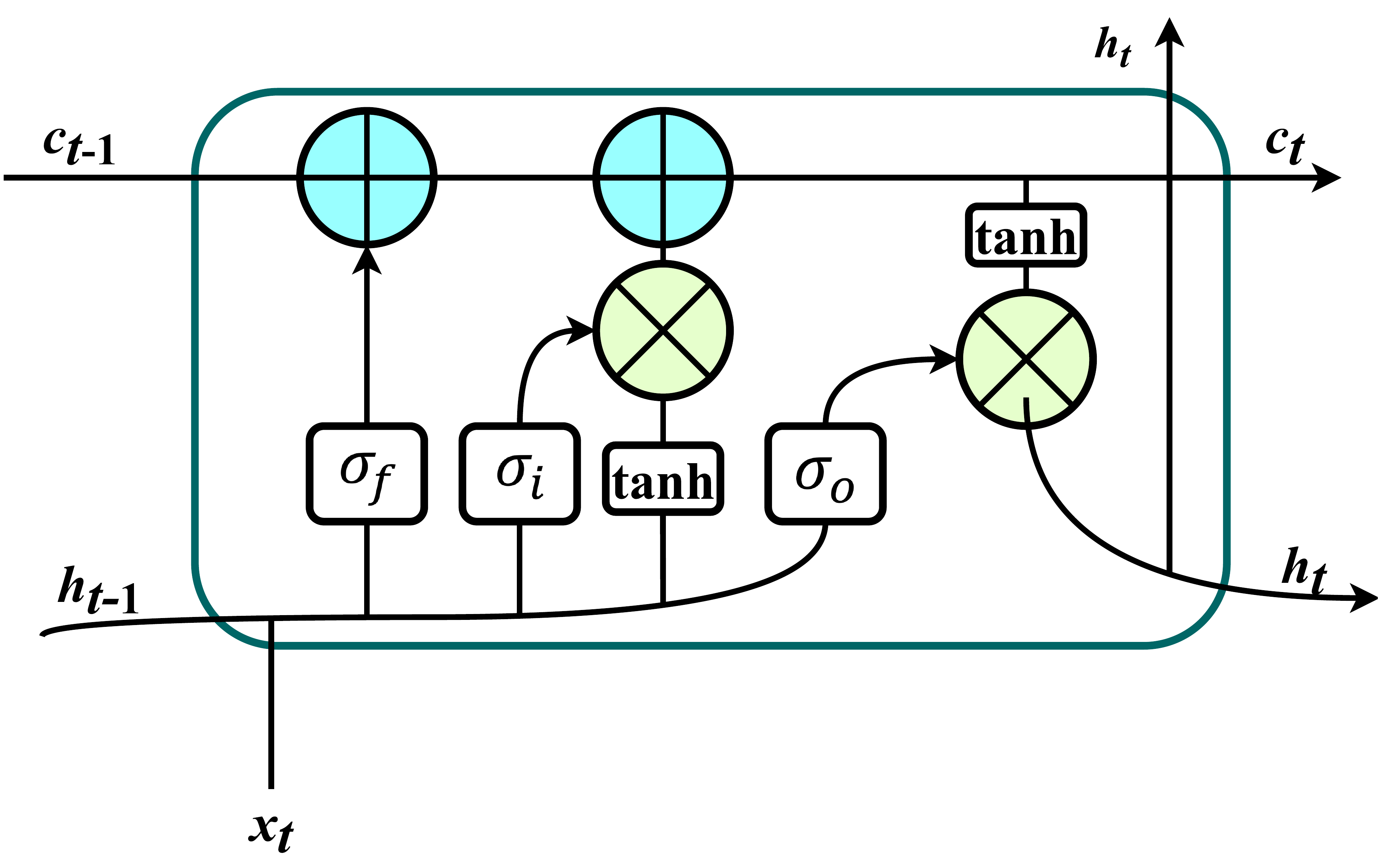}
\end{minipage}%
}%
\caption{ The subfigure (a) shows the working principle of LSTM layer, the subfigure (b) shows the working the principle of LSTM cell. LSTM is a network with a long-term memory function consisting of forgetting gate, input data, and output date, as shown in Fig.\ref{lstm}(b).}
\label{lstm}
\end{figure}
\subsection{Graph Representation}
This subsection describes the reason and process for the graph representation of optical flow and trajectories. As stated in the previous section, the model for predicting trajectories based on a single trajectory cannot solve the randomness and mutability of trajectories. To accurately predict multivariate trajectories, it is necessary to consider incorporating additional heterogeneous data into models of interaction behavior. Considering that no matter what scenario the vehicle is in, it is more concerned with moving objects. Thus the proposed method introduces data that can reflect the velocity of object motion - optical flow. The approach combines two types of heterogeneous data, optical flow, and trajectory, which respond to partial information, to accurately model the interactive behavior. In addition, previous research has demonstrated that Spatio-temporal graph convolutional networks (ST-GCN) can effectively model social behaviors; therefore, this approach employs ST-GCN for graph embedding of optical flow and trajectories.
The upper left part of Fig. \ref{framework} depicts the optical flow generation process. Previous researches have invented various algorithms for optical flow generation\cite{brox2010large,dosovitskiy2015flownet}. The paper \cite{farneback2003two} presents a novel two-frame motion estimation algorithm based on polynomial expansion transform. \cite{wu2016video} uses optical flow to represent motion and predicts dynamic visual salience by combining spatial and temporal features. The optical flow channel implements Flownet 2.0 to obtain optical flow, which can accurately reflect the instantaneous speed of objects in the view\cite{ilg2017flownet}. Under the assumption of small movement, spatial coherence\cite{fortun2015optical}, and brightness constancy, the optical flow could be computed as follows:
\begin{equation}
    \frac{\partial G }{\partial x}\Delta x + 
    \frac{\partial G }{\partial y}\Delta y + 
    \frac{\partial G }{\partial t}\Delta t = 0
\end{equation}
\begin{equation}
    \frac{\partial G }{\partial x}\Delta V_x + \frac{\partial G }{\partial y}\Delta V_y + \frac{\partial G }{\partial t} = 0
\end{equation}
where $G$ is the grayscale image, $x$ and $y$ are pixel coordinates, and $t$ represents the time index. $V_x$ and $V_y$ are  the velocity of the pixel (x,y) in the X and Y direction, respectively.
The brighter the image color is, the faster the object moves. This method emphasizes the relationship between the local features of optical flow, i.e., the relationship behind the pixels with brighter colors. To learn this relationship, the proposed method employs the ST-GCN layers for embedding optical flow graphs, efficiently reducing computational complexity.

Another type of perception data utilized in this method is the trajectory. As shown in the upper right part of Fig. \ref{framework}, the target detection module takes the original images as inputs, identifies the position of vehicles in the image, and outputs the coordinates of the top left and bottom right vertexes of the detection result bounding boxes. For $i$-th object in the image, the top left vertex is denoted as $p_{i}^{tl}=(x_{i}^{tl}, y_{i}^{tl})$, and the bottom right vertex is denoted as $p_{i}^{br}=(x_{i}^{br}, y_{i}^{br})$, $tl$ stands for top left while $br$ stands for bottom right. The sequence of points $P_{i}$ constitutes a complete trajectory, where $P_{i}=(x_i,y_i)$. Thus, the geometric center coordinates of the bounding box can be calculated as:
\begin{equation}
\left\{
\begin{aligned}
    x_{i} = \frac{x_{i}^{br} + x_{i}^{tr}}{2} \\
    y_{i} = \frac{y_{i}^{br} + y_{i}^{tr}}{2} \\
\end{aligned}
\right.
\end{equation}

The optical flow and trajectory information reflect the speed and direction of vehicles, respectively. To characterize their interactions in low illumination conditions, the proposed approach implements ST-GCN for graph representing. The method defines a new graph $G$ with attributes corresponding to the set of attributes $G_t$. In this study, the novel kernel function\cite{mohamed2020social} is adopted within the graph representing, and formulated as:
\begin{equation}
    a_{t}^{i,j} = \frac{1}{ {\| v_t^i - v_t^j \|}_2 + \epsilon}
\end{equation}
where $V = \{ v^i \ | \  \forall i \in {1,2,...,N}\}$ is the set of vertices of the graph $G_t$ as mentioned in Section  II, $\epsilon$ is an infinitesimal.

\begin{figure}[htbp]
\centering
\subfigure[]{
\begin{minipage}[t]{\linewidth}
\centering
\includegraphics[width=0.75\linewidth]{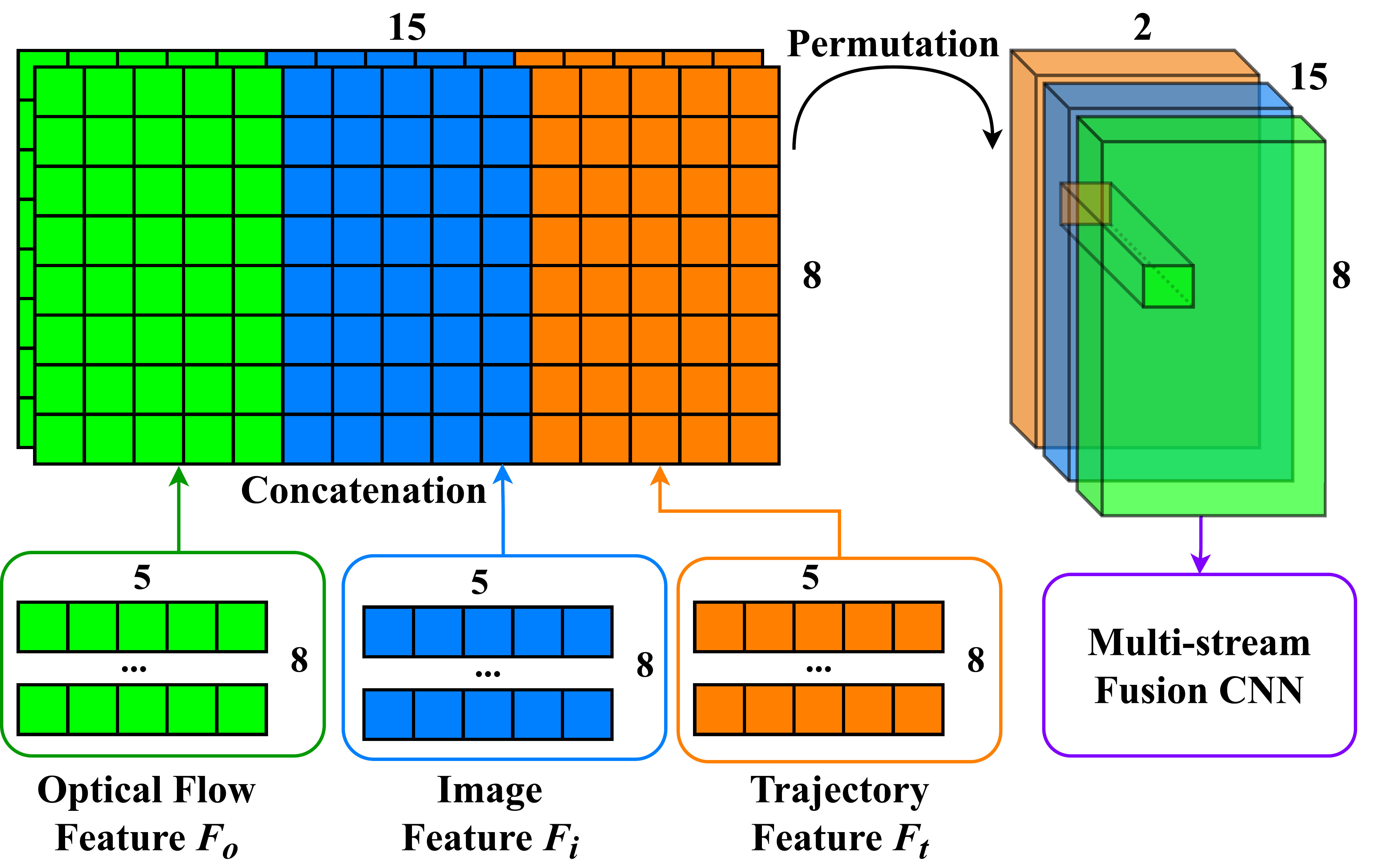}
\label{image_fusion_v1}
\end{minipage}%
}%

\subfigure[]{
\begin{minipage}[t]{\linewidth}
\centering
\includegraphics[width=0.75\linewidth]{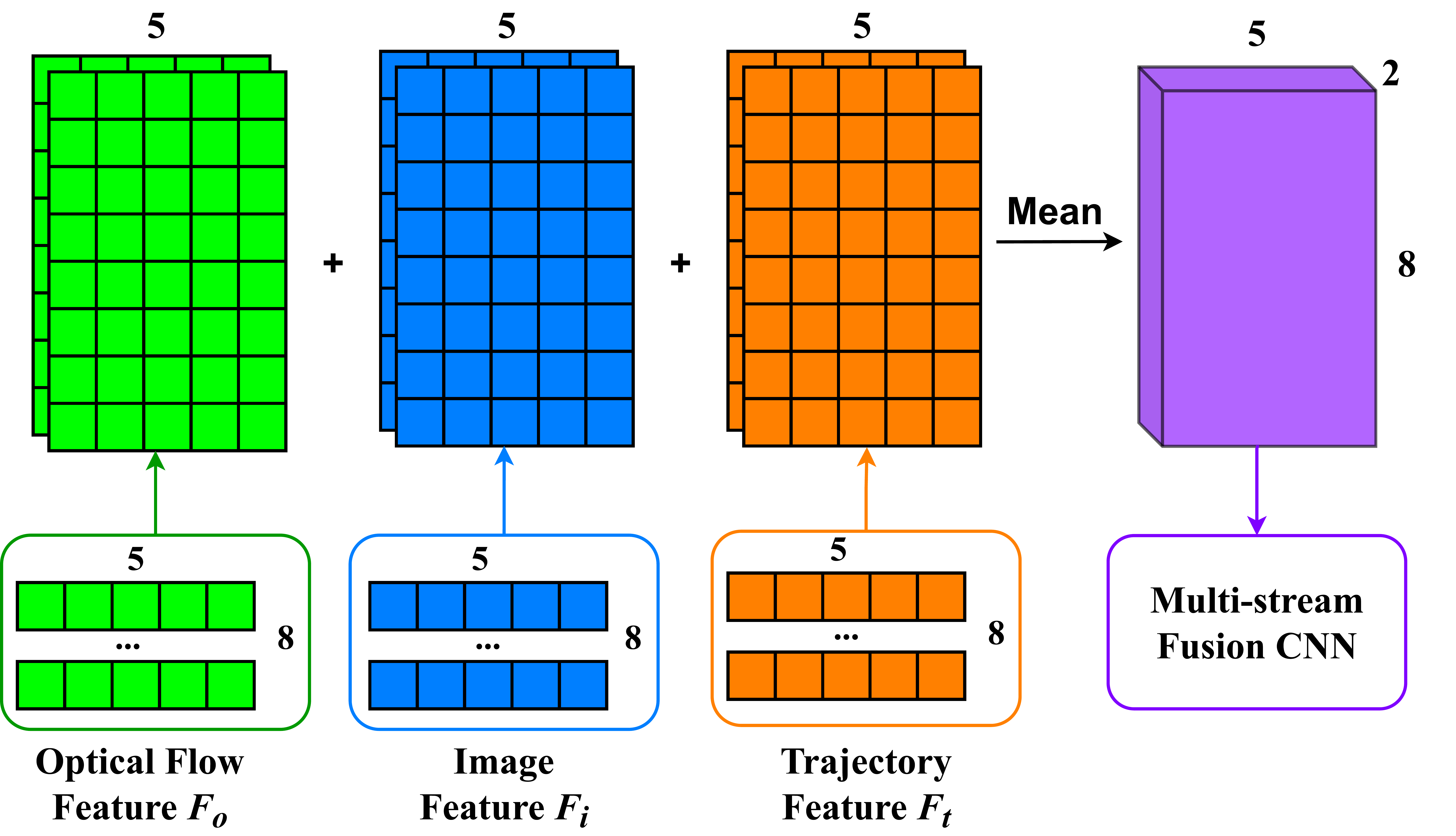}
\label{image_fusion_v2}
\end{minipage}%
}%
\caption{ The figures present two methods of feature fusion. (a) shows the stitching fusion operation (concatenation) of image feature $F_i$, optical flow feature $F_o$, trajectory feature $F_t$. (b) shows the isotopic fusion operation (calculating arithmetic mean).}
\label{fusion}
\end{figure}
After representing the interaction by using the graph, ST-GCN implements graph convolutional layers and temporal convolutional layers to extract features according to the definition given in \cite{mohamed2020social}:
\begin{equation}
     f(V^{l}, A) = \sigma (\Lambda^{-\frac{1}{2}} \hat{A}_t \Lambda^{-\frac{1}{2}}V^{l} \bm{W}^{l})
\end{equation}
where $A_t$ is symmetrically normalized by
\begin{equation}
A_t = \Lambda^{-\frac{1}{2}} \hat{A}_t  \Lambda^{-\frac{1}{2}}V^{l}
\end{equation}
\begin{equation}
\hat{A}_t = A_t + I
\end{equation}
and $\Lambda_t$ is the diagonal node degree matrix of $\hat{A}_t$, and $I$ is the identity matrix.
\subsection{Image Feature Extraction}
It is necessary to integrate image data to comprehend the features of the environment. CNN is used in the image channel to extract image features. LSTM layers are used to learn this partial and temporal information when small spatial differences between feature maps are considered. The sequence of images labeled $\bm{I}_t$ in Section II serves as input for the image feature extraction module. Firstly, images are resized to 600 × 480 before consecutively input to the convolutional neural network and LSTM layers. This process of CNN could be formulated as:
\begin{equation}
     \bm{L}_t = CNN(\bm{I}_t) 
\end{equation}
Details of the CNN network structure are shown in Fig.\ref{architecture}.
The size of the feature map output by CNN is $8 \times 15$ as shown on the left side of Fig.\ref{lstm} (a). The feature map is serialized to get a sequence with a stride of 8, and the data $1 \times 5$ from each row is used as the input of the LSTM cell.

LSTM is a network with a long-term memory function consisting of a forgetting gate, input data, and output date, as shown in Fig.\ref{lstm}(b). The forward process of LSTM could be represented as Eqs.(\ref{f})-(\ref{h}). The forget gate decides what information is left in the cell state and updates the cell state Eqs.(\ref{f})-(\ref{c1}). The input gate decides what information to discard from the cell state Eq.(\ref{i}). The output gate controls the output of the cell state Eqs.(\ref{o})-(\ref{h}). At time step $t$, the input and output of LSTM hidden layer are $x_t$ and $h_t$, and  the memory unit is $c_t$.
\begin{subequations}
\begin{align}
&f_t = \sigma(\bm{W_{xf}}x_t + \bm{W_{hf}}h_{t-1} + b_f) \label{f}\\
&c_t = f_t\odot c_{t-1} + i_t \odot tanh(\bm{W_{xc}}x_t + \bm{W_{hc}}h_{t-1} + b_c) \label{c1} \\
&i_t = \sigma(\bm{W_{xi}}x_t + \bm{W_{hi}}h_{t-1} + b_i) \label{i} \\
&o_t = \sigma(\bm{W_{xo}}x_t + \bm{W_{ho}}h_{t-1} + b_o) \label{o} \\
&h_t = o_t \odot tanh(c_t) \label{h}
\end{align}
\end{subequations}
where $\bm{W_{xf}}$, $\bm{W_{xi}}$, $\bm{W_{xo}}$ are the weight matrices in LSTM cell, $\bm{b_f}$, $\bm{b_i}$, $\bm{b_o}$ are bias, and $\sigma$ represents the Sigmoid activation function.
\subsection{Information Fusion}
This subsection first introduces components of the trajectory prediction module and the investigation of multi-stream heterogeneous data fusion methods.

The main objective of the proposed method is to predict the future trajectory of vehicles in interactive scenarios. Thus, the \emph{Trajectory Prediction Module} (TPM) is designed at the end of the framework. TPM consists of two parts: A multi-stream fusion convolutional neural network (MFC) and the trajectory prediction convolutional neural network (TPC). 

The TPC takes the optical flow graph embedding feature $\bm{F}_o$, trajectory graph embedding feature $\bm{F}_t$, and image feature $\bm{F}_i$ with the same dimensions as inputs.
Considering the structural differences between heterogeneous data from multi-stream sources, this subsection introduces two types of data fusion: stitching and isotopic fusions.

Stitching fusion can retain the feature information of multi-stream data to the maximum extent. As shown in Fig.\ref{image_fusion_v1}, stitching fusion concatenates features from three channels of the same size together and transfers them to the Multi-stream Fusion CNN, which can be formulated as Eq.(\ref{fusion_v1}).

Fig.\ref{image_fusion_v2} shows the isotopic fusion process, which performs an equal sum operation on three features with the same dimension and then takes the average. Eq.(\ref{fusion_v2}) formulates the isotopic fusion process:
\begin{equation} \label{fusion_v1}
    \bm{F}_\text{fusion} = concat(\bm{F}_t | \bm{F}_i | \bm{F}_o)
\end{equation}
\begin{equation} \label{fusion_v2}
    \bm{F}_\text{fusion} = mean(\bm{F}_t + \bm{F}_i + \bm{F}_o)
\end{equation}
\begin{equation}
    \bm{F}_\text{fusion} = MFC(\bm{F}_\text{fusion}; \bm{W}_\text{fusion})
\end{equation}
The trajectory prediction convolutional neural network (TPC), expanding the temporal dimension by using convolution, is adapted to generate final bi-variate Gaussian distributed trajectories $\bm{Y}$:
\begin{equation}
    \bm{Y} = TPC(\bm{F}_\text{fusion}; \bm{W}_\text{TP})
\end{equation}
where $\bm{W}_\text{TP}$ represents the weight matrix of TPC. Details of MFC and TPC structure are shown in Fig.\ref{architecture}.
\begin{figure*}[]
\centering
\includegraphics[width=0.85\linewidth,height=0.3\textwidth]{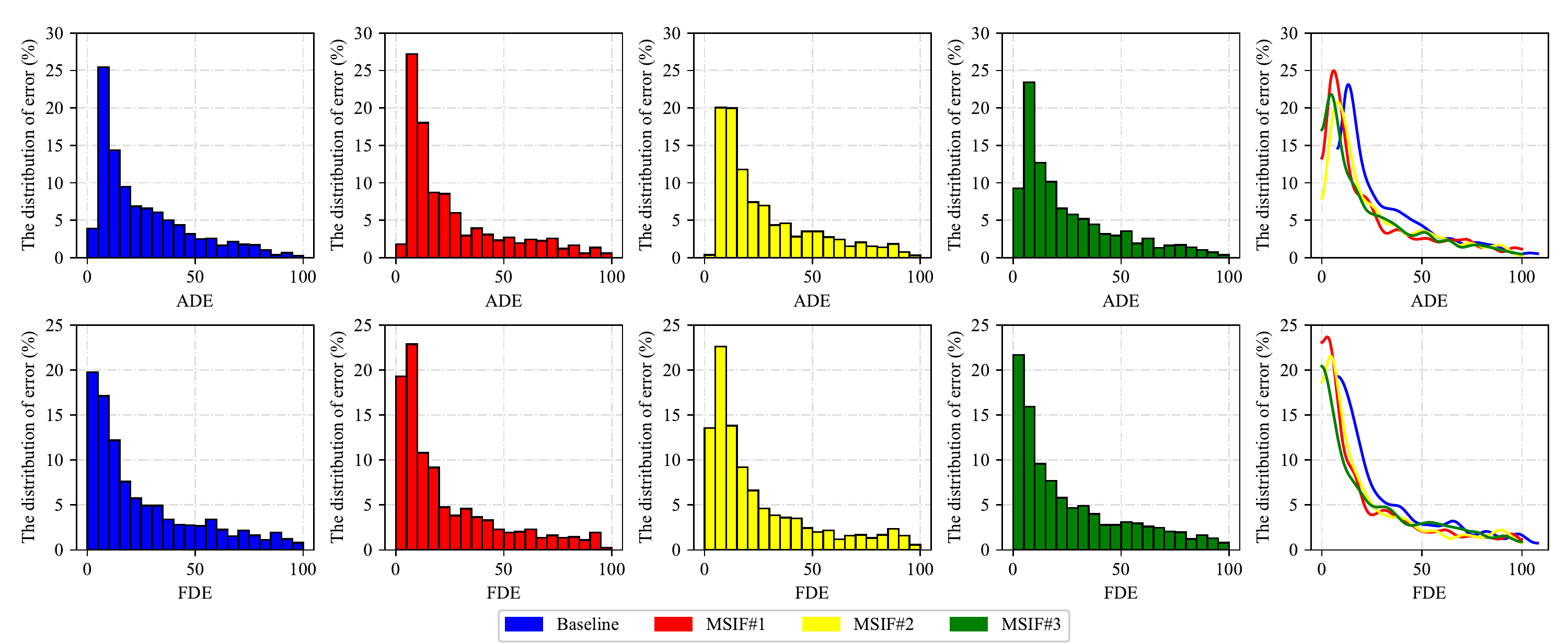}
\caption{ The histograms show comparative results of error distributions for ADE and FDE in HEV-I dataset. Error distributions of baseline, MSIF\#1, MSIF\#2, and MSIF\#3 are presented in blue, red, yellow, and green. The X-axis is the range of predicted error for each test sample. The Y-axis is the percentage of samples in a different range of errors.}
\label{hist1}
\end{figure*}
\begin{figure*}[]
\centering
\includegraphics[width=0.85\linewidth,height=0.3\textwidth]{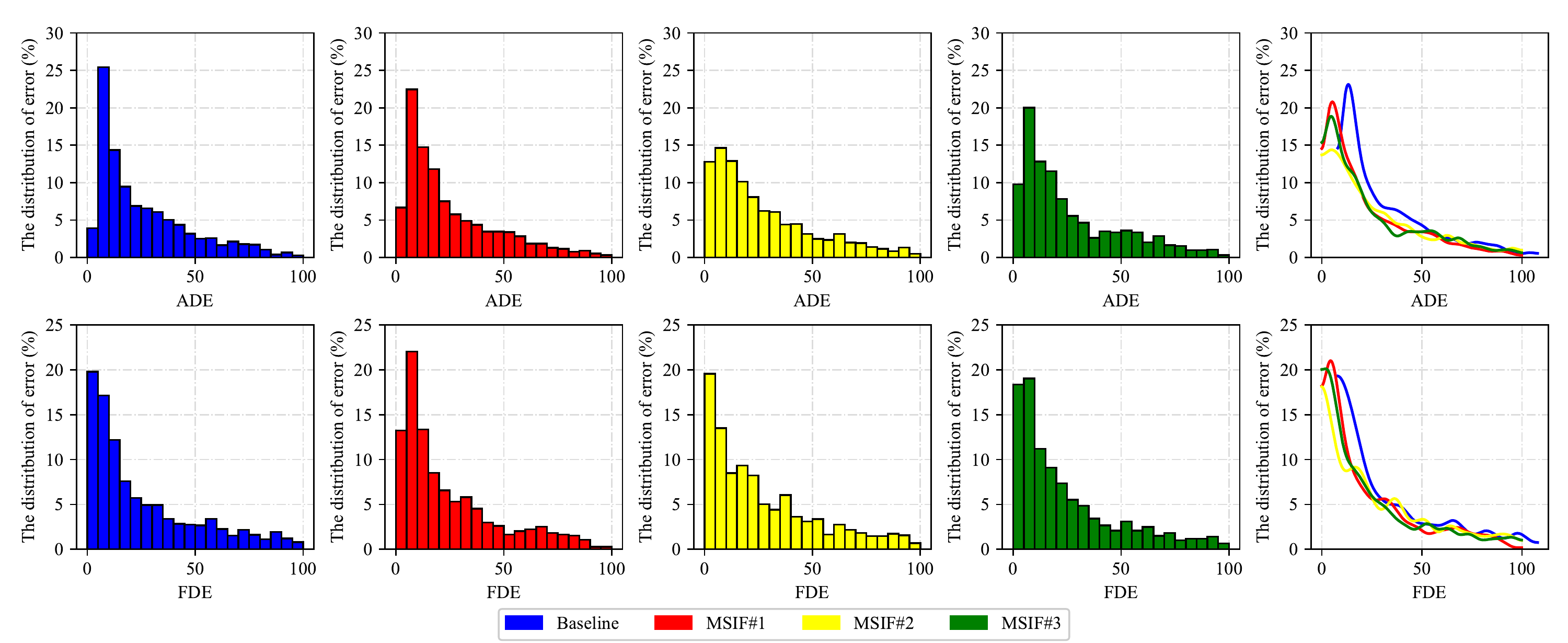}
\caption{ The histograms show comparative results of error distributions for ADE and FDE in Dark-HEV-I dataset. Error distributions of baseline, MSIF\#1, MSIF\#2, and MSIF\#3 are presented in blue, red, yellow, and green. The X-axis is the range of predicted error for each test sample. The Y-axis is the percentage of samples in a different range of errors.}
\label{hist2}
\end{figure*}
\begin{figure*}[thbp]
\centering
\includegraphics[width=0.9\linewidth]{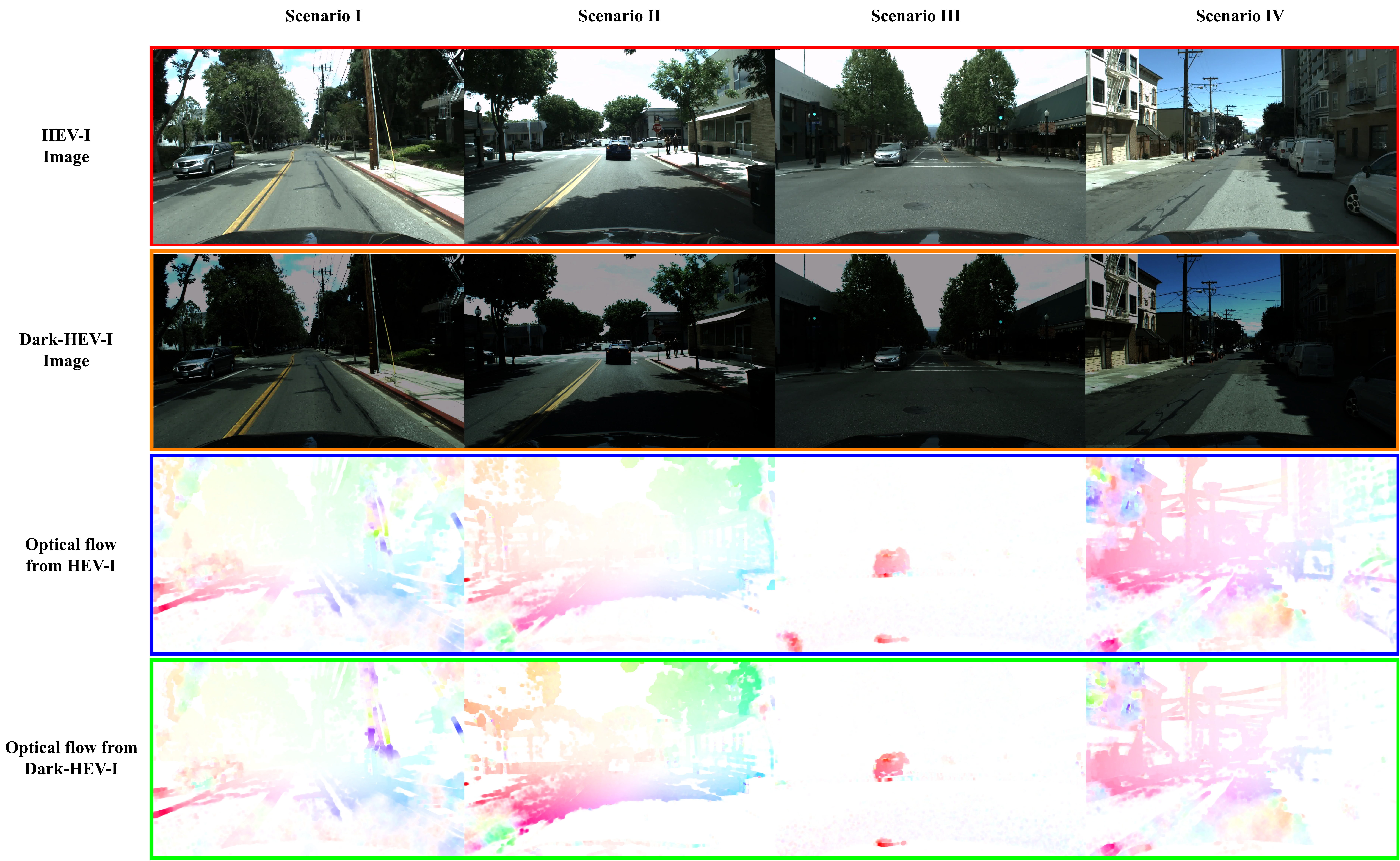}
\caption{ This figure presents the sample of HEV-I dataset and generated Dark-HEV-I dataset. \textbf{Top to bottom:} \textcolor{red}{HEV-I image}; \textcolor{orange}{Dark-HEV-I image}; \textcolor{blue}{optical flow from HEV-I image}; \textcolor{green}{optical flow from Dark-HEV-I image}. Each column corresponds to a scenario.}
\label{Dark-HEV}
\end{figure*}
\section{Experiments}\label{Experiments}
This section introduces the Honda Egocentric view-Intersection dataset, which focuses on interactive scenarios, to validate the above-mentioned method. Experiments demonstrate that the proposed method performs well on both the HEV-I and the generated Dark-HEV-I datasets. Besides, details of experiments, including evaluation metrics, baselines, implementation details, and experimental results, are presented in the following subsection. Finally, this section conducts quantitative and qualitative analyses to demonstrate the feasibility and superiority of the proposed approach.


\subsection{Dataset and Evaluation Metrics}

\subsubsection{HEV-I dataset} \ \emph{Honda Egocentric View-Intersection} (HEV-I) is a vision dataset that focuses primarily on urban intersection scenarios where vehicles move in uncertain directions due to the complexity of road layouts and traffic conditions. The reasons for selecting HEV-I in this work include: 1) Unlike other standard datasets in autonomous driving scenarios, the HEV-I dataset contains more videos and vehicles and pays close attention to the vehicle states in interactive scenarios. In contrast, a dataset such as KITTI focuses on the ego vehicle instead of interaction scenarios. For example, vehicles are parked on roadsides or drive in one direction in most of their videos. Therefore, the HEV-I dataset is preferable for trajectory prediction problems in interactive scenarios. 2) The HEV-I dataset includes driving scenes captured at various times and under varying light intensities (including backlighting and  low light), enabling the model to examine the prediction of trajectories under low illumination conditions. 3) HEV-I is an egocentric view dataset as opposed to a BEV view dataset. This vision-based dataset is more suitable for feature extracting and understanding the interaction scene of vehicles.
The HEV-I dataset contains 230 videos as 1920 × 1200 images in 10Hz and ground truth trajectories belonging to eight object classes. To obtain dense optical flow, this approach uses Flownet 2.0\cite{ilg2017flownet} with a 5x5 Region Of Interest (ROI) Pooling operator to generate a final flattened feature vector $\bm{F}_o \in \mathbb{R}^{50}$.

\subsubsection{Dark HEV-I dataset} \ The Dark-HEV-I has exactly the same scale as the HEV-I. Based on the HEV-I dataset, the new Dark-HEV-I dataset is generated to simulate a low-illumination autonomous driving environment. To simulate the low-illumination conditions, the exposure of each image is adjusted by utilizing the \emph{scikit-image} library and implement gamma correction. The principle of gamma correction could be formulated as follows\cite{poynton2012digital}:
\begin{equation} \label{gamma_correction}
    V_{out} = V_{in}^g
\end{equation}
where $V_{out}$ is the output luminance value, and $V_{in}$ is the input luminance value. $g$ denotes the \emph{gamma} value. If $\emph{gamma} > 1$, the new image will be darker than the original image. If $\emph{gamma} < 1$, the new image will be brighter than the original image. This function transforms the input image pixelwise after scaling each pixel to the range 0 to 1.
The optical flow is regenerated by the new images with low exposure. HEV-I image, Dark-HEV-I image, HEV-I optical flow, and Dark-HEV-I optical flow are presented from top to bottom in Fig.\ref{Dark-HEV}. The new images and the regenerated optical flow together form the Dark-HEV-I dataset. Each column in Fig. \ref{Dark-HEV} represents one scenario. 

\subsubsection{Metrics} \ Average Displacement Error (ADE)\cite{pellegrini2009you}, and Final Displacement Error (FDE) \cite{alahi2016social} are two metrics commonly used in trajectory prediction problems to accurately evaluate the model performance. ADE measures the average deviation from the ground truth, while FDE measures the absolute deviation at the endpoints of predicted trajectories. The lower the ADE and FDE, the better the model performance. Given that the HEV-I dataset is an image-based ego-centric  dataset, the experimental results and evaluation metrics hereinafter are calculated in pixel coordinates.

Similar to Social-STGCNN\cite{mohamed2020social} and Social-LSTM\cite{alahi2016social}, the experiments generate 20 samples based on the predicted distribution and use the following formula to compute ADE and FDE with respect to the ground truth:
\begin{equation}
     ADE =  \frac{\sum\limits_{n \in N}\sum\limits_{t \in T_e}{\left \| \hat{p}_t^n - p_t^n \right \|}_2 }{N \times T_p}
\end{equation}
\begin{equation}
     FDE =  \frac{\sum\limits_{n \in N}{\left \| \hat{p}_{t_e}^n - p_{t_e}^n \right \|}_2 }{N}
\end{equation}
where $N$ represents samples of test set, $T_p$ is the prediction time, the ground truth, and the $n^{th}$ sample coordinates at time step $t$ are denoted as $\hat{p}_{t_e}^n$ and $p_{t_e}^n$.

\
\subsubsection{Baselines}
Comparing the proposed approach with the most classical and state-of-the-art models: Structural-RNN, Social-LSTM, and Social-STGCNN, the main differences between our models and baseline are shown in the middle column of the Table \ref{table1}:
\begin{itemize}
\item {\textbf{Structural-RNN} Structural-RNN\cite{jain2016structural} takes the trajectories as input, and successfully combines the high-level representation of the Spatio-temporal graphs with the sequence learning success of recurrent neural networks.}
\item {\textbf{Social-LSTM} Social-LSTM\cite{alahi2016social} takes the trajectories as input, which models the potentially conflicting social interactions among pedestrians by adopting long short-term memory cells.}
\item {\textbf{Social-STGCNN} Social-STGCNN\cite{mohamed2020social} is a Spatio-temporal graph convolutional network that combines CNN and GCN. It extracts spatial and temporal information from the graph to generate suitable embeddings, which are then utilized by the time convolutional network to predict pedestrian trajectories.}
\item {\textbf{MSIF\#1}  The base model fuses optical and trajectory information but without image information.}
\item {\textbf{MSIF\#2}  The base model fuses image and trajectory information but without optical flow information.}
\item {\textbf{MSIF\#3}  The base model fuses image information, optical flow, and  trajectory information.}
\end{itemize}
\subsection{Implementation details and Experimental Settings}
All experiments utilize the HEV-I dataset, which is divided into training, validation, and testing sets in a ratio of 7:2:1, with the corresponding numbers being 4631:1529:665. All models are implemented using Pytorch, and experiments are run on an RTX3090 GPU. The optimizer is Adaptive Moment Estimation (Adam). The initial training rate is set as $1.0e-6$, and a learning rate scheduler is used to adjust the learning rate to its 10\% every 50 epochs. Table \ref{table2} contains a summary of the training parameters. Given the modular design of each stream, a plug-and-play method is developed to fuse heterogeneous data, satisfying the requirement for flexibility and generality for the trajectory prediction approach, so that in the future, additional types of potential perception data can be readily utilized.
\begin{table}[htbp]
\caption{Parameters of Implementation details}
\label{table2}
\begin{center}
\begin{tabular}{p{40mm}<{\centering}p{30mm}<{\centering}}
\toprule
{\textbf{Parameters}} & {\textbf{Value}}\\
\midrule 
Learning Rate & $1.0e-6$\ \\
Optimizer & Adam \ \\
Batch Size & 1024\ \\
Number of training episode & 250\ \\
Dimension of input &  2\ \\
Dimension of output & 5\ \\
Number of ST-GCN layers & 1\ \\
Number of TXPCNN layers & 5\ \\
Dimension of output & 5\ \\
Length of observed trajectory & 8\ \\
Length of predicted trajectory & 12\ \\
\bottomrule
\end{tabular}
\end{center}
\end{table}
\begin{table*}[htbp]
\caption{Quantitative results of proposed approach and baselines on HEV-I and Dark-HEV-I with metrics ADE/FDE}
\label{table1}
\begin{center}
\begin{tabular}{p{25mm}<{\centering}p{25mm}<{\centering}p{25mm}<{\centering}p{25mm}<{\centering}p{25mm}<{\centering}p{25mm}<{\centering}}
\toprule
{\ul}&\multicolumn{3}{c}{\textbf{Multi-stream}}&{\textbf{HEV-I}}&{\textbf{Dark-HEV-I}}\\
\textbf{Model}&\textbf{Trajectory}&\textbf{Optical flow}&\textbf{Image}&\textbf{ADE\ / \ FDE\ } & \textbf{ADE\ / \ FDE\ }\\
\midrule 
Structural-RNN&$\checkmark$&$\times$&$\times$&35.40\ / \ 53.99&-\ / \ -\ \\
Social LSTM&$\checkmark$&$\times$&$\times$&\textbf{32.57}\ / \ 51.23&-\ / \ -\ \\
Social-STGCNN&$\checkmark$&$\times$&$\times$&58.65\ / \ 56.95&-\ / \ -\ \\
MSIF\textbf{\#1}&$\checkmark$&$\checkmark$&$\times$&51.10\  / \ \textbf{50.76}&50.32\ / \ \textbf{50.15}\\
MSIF\textbf{\#2}&$\checkmark$&$\times$&$\checkmark$&60.95\ / \ 52.93 &220.60\ / \ 140.52\\
MSIF\textbf{\#3}&$\checkmark$&$\checkmark$&$\checkmark$& \textbf{33.18}\ / \ \textbf{45.77} &\textbf{44.94}\ / \ 64.76\\
\bottomrule
\end{tabular}
\end{center}
\end{table*}
\subsection{Experiment \uppercase\expandafter{\romannumeral1}: Trajectory Prediction in Normal Scenarios}
This study conducts experiments on heterogeneous multi-stream sensing data in HEV-I dataset. Using Structural-RNN, Social-LSTM, and Social-STGCNN as the baseline, this experiment investigates the impact of scenario images and optical information on the accuracy of intersection trajectory prediction in standard illumination scenarios. As shown in Table \ref{table1}, baselines only utilize trajectories of ground truth bounding boxes, based on which the proposed approach incorporates optical flow and image information. In this comparative study, the difference between MSIF\#1 and MSIF\#2 is that the input of MSIF\#1 is trajectories and optical, whereas the input of MSIF\#2 is trajectories and images, while the MSIF\#3 combines the information of the three.
\begin{figure*}[ht]
\centering
\includegraphics[width=\linewidth]{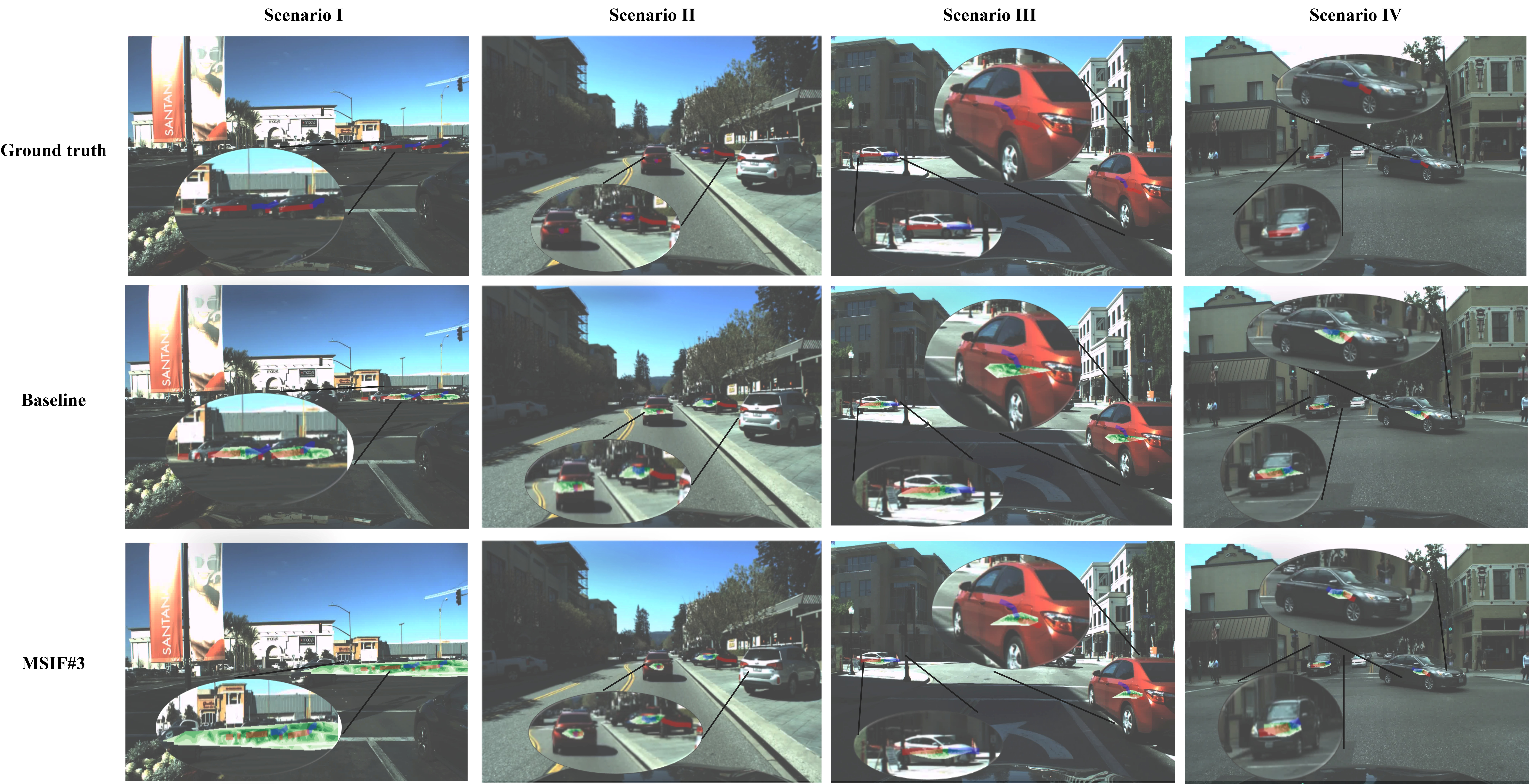}
\caption{ This figure shows the performance of the proposed approach in the low-illumination scenario. We choose four illumination scenarios in the test set and visualize the \textcolor{green}{predicted trajectories distribution}. \textbf{Top to bottom}: Ground truth, Baseline, MSIF\#3. The result shows that trajectories distribution of MSIF\#3 is closer to the ground truth.}
\label{darkCondition}
\end{figure*}
The experimental results presented in Table \ref{table1} demonstrate that the proposed method outperforms the baseline Social-STGCNN in HEV-I dataset. Comparing the baseline to MSIF\#1, ADE decreases from 58.65 px to 51.10 px, and FDE decreases from 56.95 px to 50.76 px, indicating that optical flow improves the accuracy of trajectory prediction results. MSIF\#2 achieves ADE of 60.95 px and FDE of 52.93 px, which indicates that MSIF\#2 (with images) is mediocre on the metrics of ADE and FDE but does not indicate its misunderstanding of interactive scenarios adequately. The best performing model MSIF\#3 achieves ADE of 33.18 px and FDE of 45.77 px, representing respective increases of 43.43\% and 19.63\% over the baseline.

Fig.\ref{hist1} presents comparative results of the error distribution. Baseline, MSIF\#1, MSIF\#2, and MSIF\#3 error distributions are depicted in blue, red, yellow, and green, respectively. The X-axis depicts the predicted error range for each test sample. The Y-axis represents the percentage of samples within each error range. The closer the histogram is to the Y-axis, the more accurate the prediction. In Fig.\ref{hist1}, the error percentages of ADE and FDE in the range of 0 px to 5 px for MSIF\#3 (green) are more than 20\%, far exceeding those of MSIF\#1. For MSIF\#1, the proportion of ADE between 8 and 12 is close to 35\%, and the percentage of FDE between 5 px and 10 px is close to 30\%. The closer the histogram of MSIF\#3 is to the Y-axis, the higher the percentage, so MSIF\#3 has the best prediction performance.

\begin{figure}[htbp]
\centering
\includegraphics[width=0.9\linewidth]{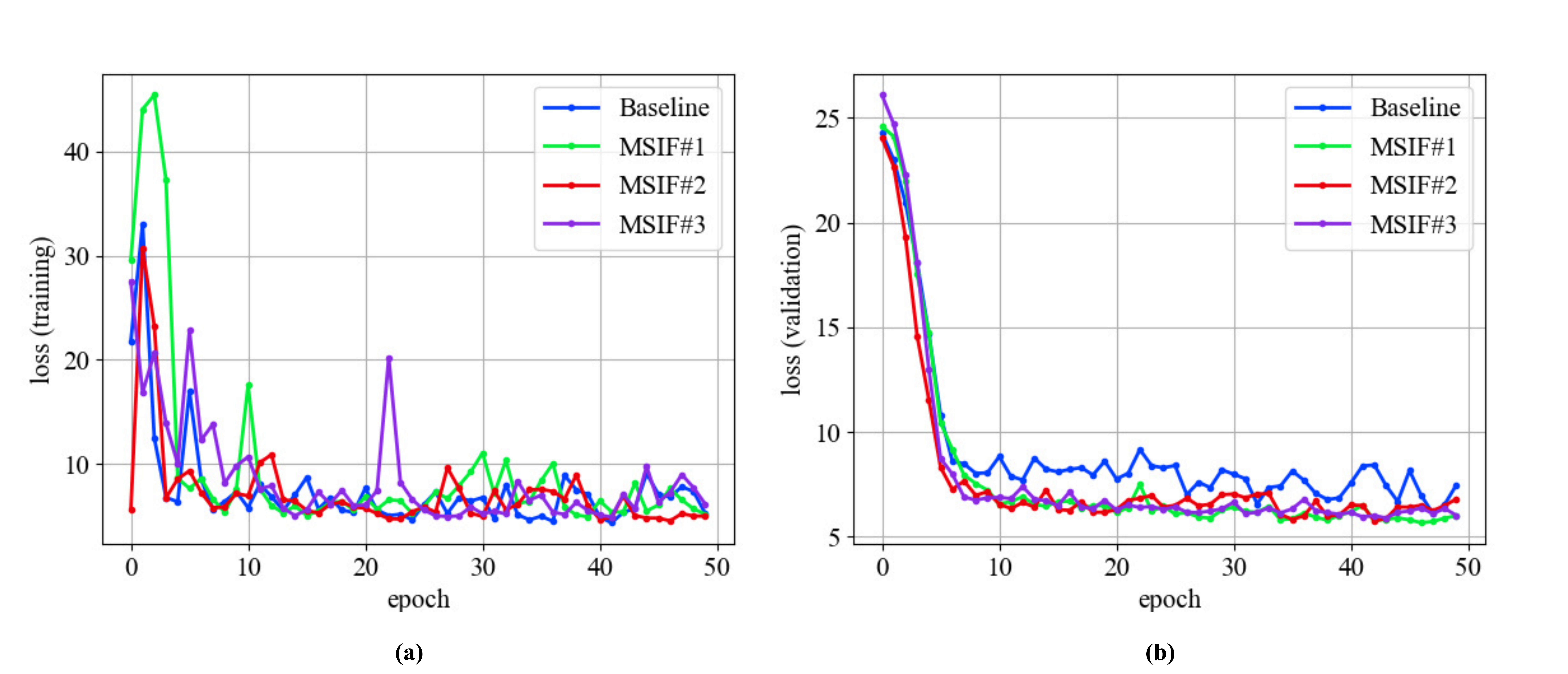}
\caption{ Figure (a) reflects the loss during training and figure (b) reflects validation process. The loss value during training is basically stable after 20 epochs, and the loss value during validation decreases smoothly in the early stage and stops decreasing after 10 epochs. The loss curve demonstrates that the model is able to fit the trajectory effectively.}
\label{loss}
\end{figure}

Fig.\ref{loss} reflects the loss during the training and validation process. The loss value during training is stable after 20 epochs, whereas the loss value during validation decreases gradually in the early stages and stops decreasing after ten epochs. The loss curve demonstrates that the model can effectively fit the trajectory. 
After feature extraction by the image channel, the scene's high-dimensional features and minor spatial differences at different time steps are reflected in the scene's feature map.
\begin{figure}[htbp]
\centering
\includegraphics[width=0.9\linewidth,height=0.5\textwidth]{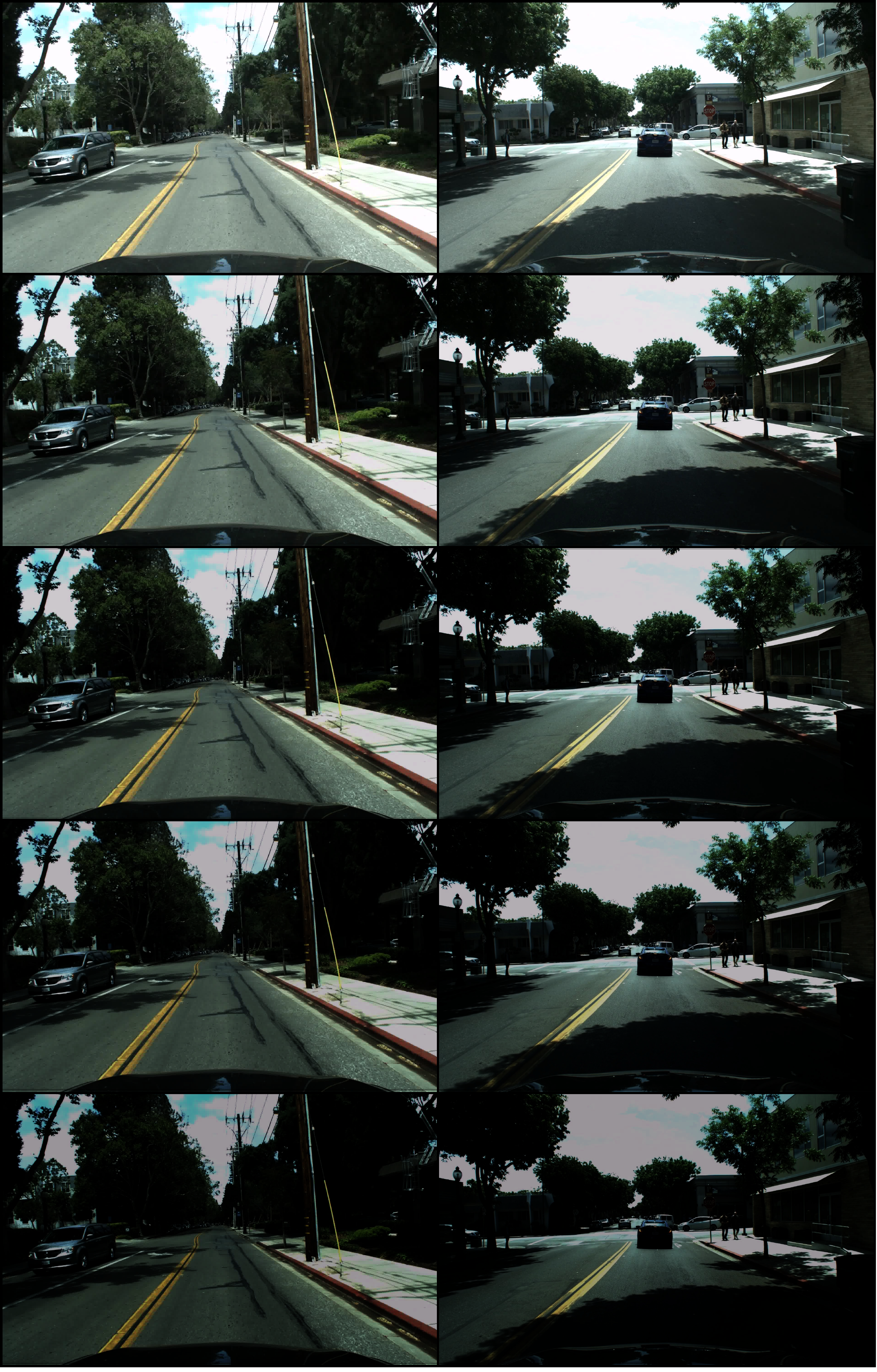}
\caption{ This figure shows the gradual change in illumination for two scenes. \textbf{Top to bottom:} gamma is equal to 1, 1.4, 1.8, 2.0, and 2.5, respectively.}
\label{illuminationChange}
\end{figure}
\subsection{Experiment \uppercase\expandafter{\romannumeral2}: Trajectory Prediction in Low-illumination Scenarios}
This experiment is conducted to validate the performance of the proposed method in low-illumination scenarios. The experimental results of the generated Dark-HEV-I dataset presented in Table \ref{table1} demonstrate that the proposed MSIF  method achieves accurate prediction results under low-illumination conditions. MSIF\#1 has an ADE of 50.32 px and an FDE of 50.15 px. MSIF\#2 achieves ADE of 220.60 px and FDE of 140.52 px, indicating that only integrating images with low illumination significantly affects the comprehension of scenarios. MSIF\#3 achieves ADE of 44.94 px and FDE of 64.76 px. MSIF\#1 achieves a lower FDE, while MSIF\#3 has a lower ADE in the Dark-HEV-I dataset. Fig.\ref{hist2} presents comparative results of the error distribution in the Dark-HEV-I dataset. MSIF\#2 (yellow) is underperforming, as the ADE is lower than 15\% in each interval in the range of 0 px to 100 px. In terms of ADE, the performance of MSIF\#3 (green) is similar to MSIF\#1 (red) with 20\% the percentage of the ADE error distribution. While in terms of FDE, the percentage of the error distribution of FDE is greater than 15\% in both the range of 0 px to 5px and 5px to 10 px. Furthermore, the distribution curve (the 5th column) shows that the histograms of MSIF\#1 and MSIF\#3 are closer to the Y-axis which indicates that MSIF\#1 and MSIF\#3 outperform the baseline in low illumination conditions.

The performance of the proposed method in the low-illumination scenario is depicted in Fig. \ref{darkCondition}. We select four poor illumination scenarios from the Dark-HEV-I dataset and visualize the distribution of predicted trajectories. The ground truth, the result of baseline, and the result of MSIF\#3 are presented from top to bottom. Although scenario I is a complex scenario where many vehicles pass through the intersection at fast speed, the predicted trajectory of MSIF\#3 effectively covers the ground truth, which demonstrates the accuracy of MSIF\#3 for long trajectory prediction. Scenarios II is a car-following case, in which the predicted result of MSIF\#3 is more reasonable than the baseline. In scenario III and scenario IV, MSIF\#3 can forecast the direction of the opposite vehicles. In summary, the predicted distribution of MSIF\#3 trajectories is closer to the ground truth, indicating that MSIF\#3 achieves accurate trajectory prediction in a low-brightness environment.

\subsection{Experiment \uppercase\expandafter{\romannumeral3}: Validation of Adaptability in different Illumination Scenarios}
The first and the second experiment validate the performance of the proposed method in standard and low-illumination conditions, respectively. This experiment is conducted to verify the adaptability of the proposed MSIF to different illumination conditions. The illumination of the scenario is changed by using the gamma correction as mentioned in Eq.\ref{gamma_correction}, which adjusts the exposure degree (gamma from 1.0 to 2.5) of the images to simulate the different light levels throughout the day. Fig. \ref{illuminationChange} shows the gradual change in illumination for two scenes. From top to bottom, the \emph{gamma} value is equal to 1, 1.4, 1.8, 2.0, and 2.5, respectively. Then, these newly generated images are used to produce corresponding optical flow information.

This experiment uses the model weights obtained by training at the Dark-HEV-I dataset (gamma equal to 2) to test the trajectory prediction performance under other light intensities. As shown in Table \ref{table4}, when gamma varies from 1.0 to 2.5, both the ADE and FDE metrics of MSIF\#1 fluctuate in a small range, and metrics of MSIF\#3 show an upward trend. As MSIF\#3 integrates the image information with low light level, but MSIF\#1 does not, it is reasonable to speculate that the low-illumination images do have a serious impact on the trajectory prediction results. 
\begin{table}[htbp]
\caption{Comparative results with Different Illumination Conditions on MSIF\#1 and MSIF\#3}
\label{table4}
\begin{center}
\begin{tabular}{p{20mm}<{\centering} p{25mm}<{\centering} p{25mm}<{\centering}}
\toprule
\multirow{2}{*}{\textbf{Gamma}} & {\textbf{MSIF\#1}} & {\textbf{MSIF\#3}} \\
 & {\textbf{ADE\ / \ FDE}}  & {\textbf{ADE\ / \ FDE}}\\
\midrule 
\textbf{1.0} & 51.10\ / \ 50.76  & 33.18\ / \ 45.47\ \\
1.4 & 51.41\ / \ 49.28 & 38.55\ / \ 52.68\ \\
1.8 & 50.84\ / \ 51.26 & 45.27\ / \ 49.32\ \\
\textbf{2.0} & \ 50.32\ / \ 50.15 & 44.94\ / \ 64.76\ \\
2.5 & 79.24\ / \ 88.07 & 67.95\ / \ 80.41\ \\
\bottomrule
\end{tabular}
\end{center}
\end{table}
\subsection{Experiment \uppercase\expandafter{\romannumeral4}: Ablation Study}
This subsection investigates various feature fusion techniques in an effort to identify a suitable method for producing the most accurate predictions. As shown in Fig.\ref{architecture}, the proposed approach adapts three different fusion methods (Mean, Weighted Mean, and Concatenation). Referring to Fig.\ref{architecture}, the output features of all the three channels have the same dimension, such as [1, 5, 8, 2]. The fusion module calculates the numerical mean or concatenates these features in the second dimension. To ensure that the output of the fusion operation meets the TPC, the fused feature will be input MFC to change the number of channels if concatenation fusion is selected.
\begin{table}[htbp]
\caption{Quantitative results of MSIF\#3 with different Fusion Methods on HEV-I and Dark-HEV-I}
\label{table3}
\begin{center}
\begin{tabular}{p{20mm}<{\centering} p{25mm}<{\centering} p{25mm}<{\centering}}
\toprule
\multirow{2}{*}{\textbf{Fusion Method}} & {\textbf{HEV-I}} & {\textbf{Dark-HEV-I}} \\
 & {\textbf{ADE\ / \ FDE}}  & {\textbf{ADE\ / \ FDE}}\\
\midrule 
Mean & 60.28\ / \ 54.30  & 915.42\ / \ 972.13\ \\
Weighted Mean & 57.23\ / \ 52.40 & 168.33\ / \ 77.02\ \\
FCNN ($\times$1) & 88.69\ / \ 64.50 & 107.27\ / \ 82.21\ \\
FCNN ($\times$2) & \textbf{33.18}\ / \ \textbf{45.77} & \textbf{44.94}\ / \ 64.76\ \\
FCNN ($\times$3) & 65.09\ / \ 60.19 & 79.44\ / \ 68.18\ \\
\bottomrule
\end{tabular}
\end{center}
\end{table}
As shown in Table \ref{table3}, the feature concatenation methods outperform other fusion methods. The TPC with two convolutional neural network layers (FCNN x2) achieve the lowest ADE and FDE in both HEV-I and Dark-HEV-I datasets. The Mean feature fusion method get ADE of 60.28 and FDE of 54.30, however, due to the low illumination conditions in Dark-HEV-I, the Mean fusion method does not perform well, with ADE of 915.43 and FDE of 972.13. FCNN (x2) are finally implemented in the TPC module due to its better performance.

\section{Conclusion} \label{Conclusion}
To address trajectory prediction issue in low-light conditions, this article proposes MSIF, a multi-stream information fusion-based approach, considering the vehicles interaction in low-illumination environment. The image channel uses a convolutional neural network and LSTM layers for feature extraction and scene understanding. ST-GCN describes the interactivity of vehicles in both the optical flow channel and the trajectory channel. The proposed approach uses the Trajectory Prediction Module (TPM) to achieve feature fusion and trajectory prediction. To simulate the low-illumination conditions, the Dark-HEV-I dataset is created. The model is validated using both the HEV-I dataset and the generated Dark-HEV-I dataset. The multi-stream comparative experiment demonstrates that the proposed method outperforms baseline methods in terms of trajectory prediction metrics. The study on feature fusion demonstrates that our method effectively combines multi-stream heterogeneous data. The qualitative analysis demonstrates that the trajectories predicted by our model in complex interaction scenarios are more reasonable and realistic, demonstrating the adaptation to a low-light environment and the achievement of scene understanding. This method is applicable to intelligent networked vehicle driving scenarios and can be implemented on the roadside for a variety of applications. 

In the future, we will investigate the application of richer perception data for trajectory predictions in extreme conditions, as well as more effective data fusion techniques.

\ifCLASSOPTIONcaptionsoff
  \newpage
\fi



%


\section{Appendix}

\begin{figure}[th]
\centering
\includegraphics[height=0.7\textwidth]{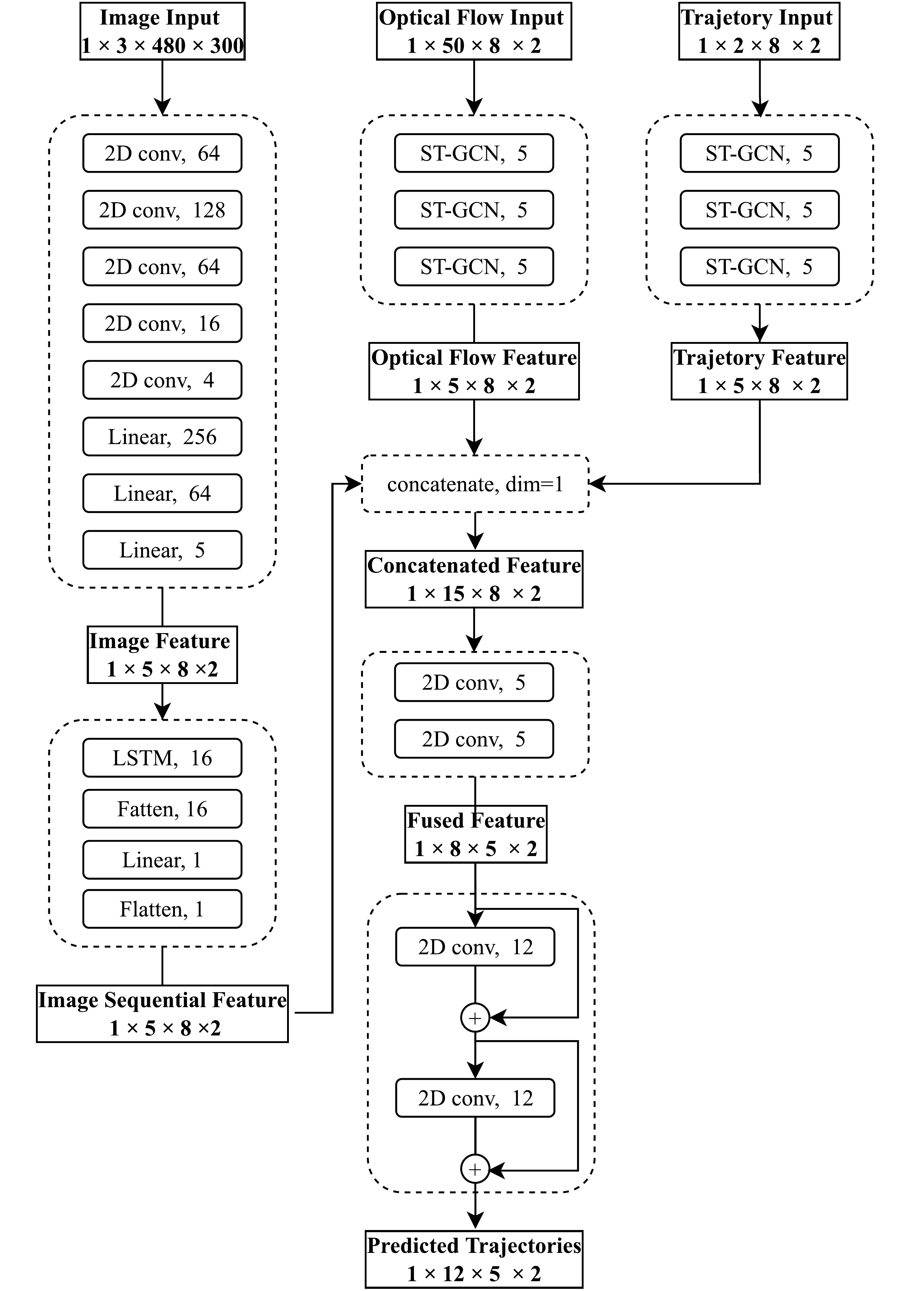}
\caption{ This figure presents the detail architecture of MSIF, including the image channel, the optical channel and the trajectory channel. The upper left side shows the architecture details of image feature extraction. The upper middle and upper right parts show layer details of optical flow channel and trajectory channel, which consist of Spatio-temporal GCN (ST-GCN). The lower middle part shows how features from three channels fusing and architecture details of multi-stream Fusion CNN. }
\label{architecture}
\end{figure}

Fig.\ref{architecture} presents the detailed architecture of MSIF, including the image channel, optical channel, and trajectory channel. The upper left side of the Fig.\ref{architecture} shows architecture details of image feature extraction. The upper middle and upper right parts show layer details of the optical flow channel and trajectory channel, which consist of Spatio-temporal GCN (ST-GCN). The lower middle part shows how features from three channels fuse and the architecture details of multi-stream Fusion CNN. 

\small
\bibliographystyle{IEEEtran}
\bibliography{bare_jrnl.bib}
\end{document}